\documentclass[runningheads]{llncs}
\usepackage[T1]{fontenc}
\usepackage{graphicx}

\usepackage{amsfonts}
\usepackage{booktabs}
\usepackage{multirow}
\usepackage{makecell}
\usepackage{graphicx}
\usepackage{subcaption}
\usepackage[font=small]{caption}

\newcommand{\coolname}{\textsc{Ilov3Splat}}

\usepackage{xcolor}
\usepackage[colorlinks=true,
            allcolors=black,
            urlcolor=blue]{hyperref}

\usepackage[acronym]{glossaries}

\newacronym{nerf}{NeRF}{Neural Radiance Field}
\newacronym{gs}{3D-GS}{3D Gaussian Splatting}
\newacronym{mhe}{MHE}{Multi-resolution hash encoding}
\newglossaryentry{ils}{
  name={Ilov3Splat},
  description={Instance-Level Open-Vocabulary 3D scene understanding built on Gaussian Splatting}
}
\glsdisablehyper

\begin{document}
\title{\textsc{Ilov3Splat}: Instance-Level Open-Vocabulary \\3D Scene Understanding in Gaussian Splatting
}
\titlerunning{\textsc{Ilov3Splat}: Instance-Level Open-Vocabulary 3D-GS}
\author{Binh Long Nguyen\inst{1,2} \and
Kien Nguyen\inst{1} \and
Sridha Sridharan\inst{1} \and
Clinton Fookes\inst{1} \and
Peyman Moghadam \inst{1,2}
}

\authorrunning{Long Nguyen et al.}
\institute{School of Electrical Engineering and Robotics,
Queensland University of Technology (QUT), Brisbane, QLD 4000, Australia
\email{\{binhlong.nguyen, k.nguyenthanh, s.sridharan, c.fookes, peyman.moghadam\}@qut.edu.au}\\
\and
CSIRO Robotics, CSIRO, Brisbane, QLD 4069, Australia\\
\email{\{binhlong.nguyen, peyman.moghadam\}@csiro.au}}
\maketitle              %
\begin{abstract}
We introduce \coolname{}, a novel framework for instance-level open-vocabulary 3D scene understanding built on 3D Gaussian Splatting (3D-GS). Most prior work depends on 2D rendering-based matching or point-level semantic association, which undermines cross-view consistency, lacks coherent instance-level reasoning, and limits precision in downstream 3D tasks.
To address these limitations, our method jointly optimizes scene geometry and semantic representations by augmenting Gaussian splats with view-consistent feature fields. Specifically, we leverage multi-resolution hash embedding to efficiently encode language-aligned CLIP features, enabling dense and coherent language grounding in 3D space. We further train an instance feature field using contrastive loss over SAM masks, supporting fine-grained object distinction across views. At inference time, CLIP-encoded queries are matched against the learned features, followed by two-stage 3D clustering to retrieve relevant Gaussian groups. 
This enables our framework to identify arbitrary objects in 3D scenes based on natural language descriptions, without requiring category supervision or manual annotations. Experiments on standard benchmarks demonstrate that \coolname{} outperforms prior open-vocabulary 3D-GS methods in both object selection and instance segmentation, offering a flexible and accurate solution for language-driven 3D scene understanding.
Project page: \url{https://csiro-robotics.github.io/Ilov3Splat}.

\end{abstract}

\section{Introduction}
\label{sec:intro}

Bridging the gap between raw 3D perception and semantic understanding is a fundamental challenge for intelligent systems that must reason about, navigate, and manipulate the physical world. A comprehensive 3D scene understanding system should go beyond geometric reconstruction to support object-level reasoning, open-ended language queries, and downstream tasks such as navigation and manipulation~\cite{vidanapathirana2025wildscenes}. This requires a unified framework that is accurate, efficient, and capable of grounding semantic perception in 3D space while maintaining consistency across views.

A promising direction is open-vocabulary 3D scene understanding, which enables models to identify arbitrary objects in 3D scenes based solely on natural language descriptions, removing the constraints of predefined category sets and manual annotations. Early approaches operated on point clouds~\cite{choe2022pointmixer,zhang2022pointclip}, but despite their simplicity, they suffer from sparsity and lack the fine-grained structure needed for precise object-level reasoning~\cite{yang2024tulip}. To overcome this, 
subsequent methods~\cite{kerr2023lerf,liu2023weakly} adopted 
\gls{nerf}~\cite{mildenhall2021nerf}, which offers dense volumetric reconstructions and has demonstrated strong capabilities in tasks such as object-level registration~\cite{hausler2024reg}, semantic scene 
understanding~\cite{zhi2021place}, open-vocabulary object 
retrieval~\cite{kerr2023lerf}, and instance-level 
segmentation~\cite{liu2023weakly}. However, \gls{nerf} methods are hindered by slow rendering and weak semantic interpretability. \gls{gs}~\cite{kerbl20233d}  overcomes these drawbacks through differentiable rasterization of explicit 3D primitives, achieving superior speed and fidelity, and has quickly become the backbone of choice for dense semantic scene understanding~\cite{zhou2024feature,ye2024gaussian,choi2024click,qin2024langsplat}. Nevertheless, instance-level reasoning within 
\gls{gs} remains an open challenge.

\begin{figure}[t]
	\centering
	\includegraphics[width=0.9\columnwidth]{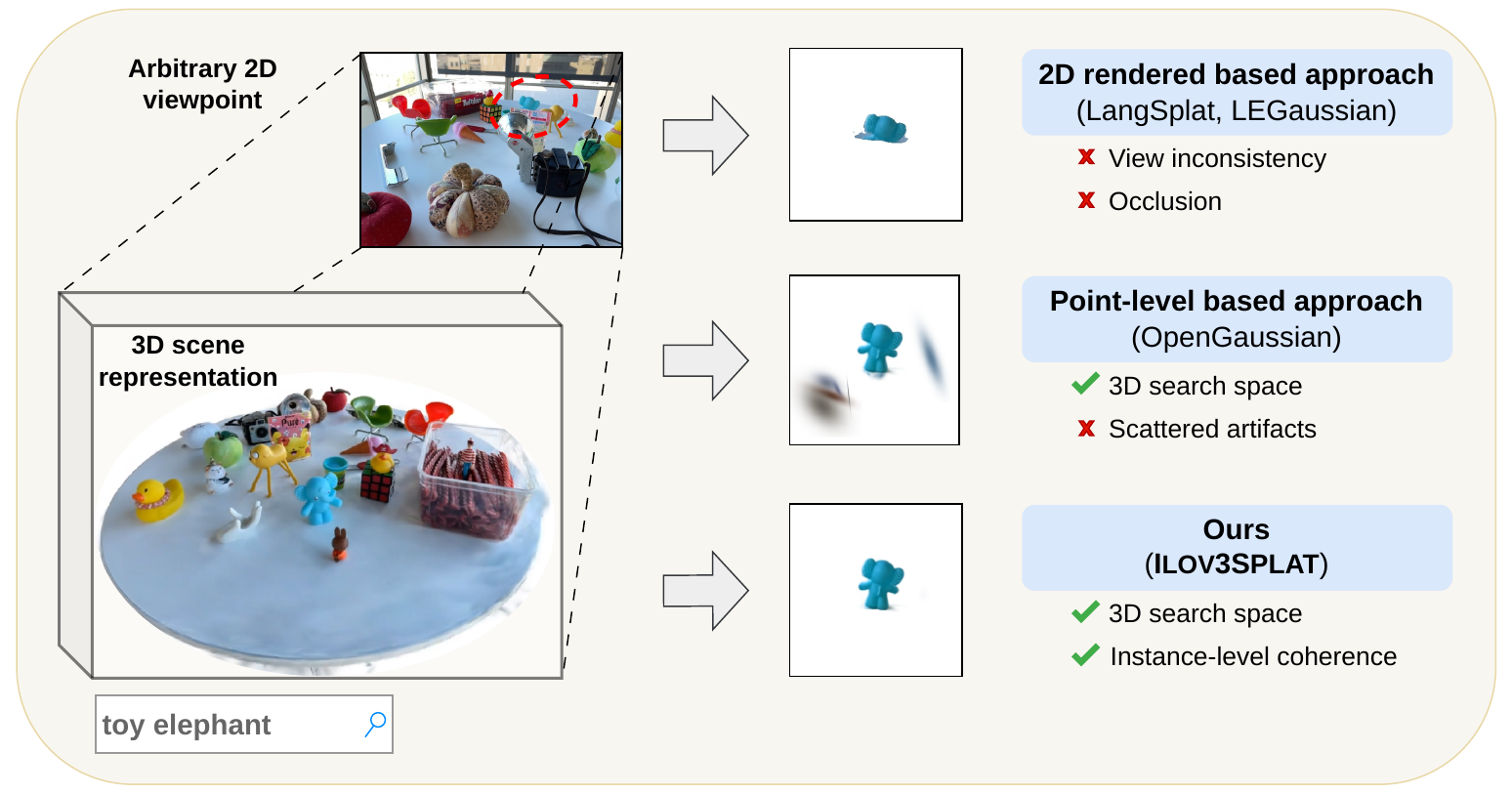}
	\centering
	\caption{Comparison between 2D-rendered, point-level, and our instance-level 3D open-vocabulary understanding approaches.
    Prior methods either rely on rendered 2D supervision, which lacks cross-view consistency, or point-level feature matching, which often results in scattered or noisy activations. In contrast, our method introduces a unified instance-level representation that maintains semantic coherence across views and enables precise retrieval of object-centric 3D Gaussians via text queries.}
	\label{fig:comparison}
\end{figure} 

Existing \gls{gs}-based approaches to open-vocabulary 3D scene understanding face several key limitations that hinder precise instance-level reasoning (refer to Figure~\ref{fig:comparison}). First, many methods rely on  2D rendering-based supervision or compressed scene-level embeddings, such as  LangSplat~\cite{qin2024langsplat} and LEGaussian~\cite{shi2024language}, which can introduce inconsistencies across views and limit spatial precision. Since features are typically derived from rendered 2D projections, they struggle to capture occluded or partially visible geometry, undermining the inherent 3D representational strengths of Gaussian-based models. Second, the frequent decoupling of geometry reconstruction and semantic learning into separate stages~\cite{qin2024langsplat,zuo2025fmgs} can lead to semantic drift and hinder joint optimization. Third, several pipelines rely on complex, scene-specific preprocessing to determine the number of instances required for 2D cross-view matching or fixed-size codebooks~\cite{qin2024langsplat,ji2025fastlgs,wu2024opengaussian}, reducing generalizability to novel scenes and object types. Finally, instance-level understanding remains underexplored, particularly the explicit separation of multiple object instances belonging to the same semantic category. 
This limitation is reflected in current methods, which often associate semantics independently with individual Gaussian points, as in OpenGaussian~\cite{wu2024opengaussian}, thereby limiting representations to point-level semantic referring.
Without instance-aware mechanisms, semantic responses may bleed into the background or unrelated regions, degrading the precision of object localization. By contrast, instance-level representations enable object-centric reasoning that is essential for downstream tasks such as 3D instance segmentation, object-level querying, and interaction~\cite{choi2024click,ye2024gaussian}. 

To overcome these challenges, we introduce \coolname{}, a 
unified framework for instance-level open-vocabulary 3D scene 
understanding built on \gls{gs}. Our method jointly learns geometry and semantics by embedding each Gaussian with compact, view-consistent feature fields that encode both language-aligned and instance-discriminative signals. Specifically, we design a 
multi-resolution hash embedding to efficiently represent CLIP-aligned features in 3D, and train a separate instance feature field using contrastive learning on multi-view SAM masks to support fine-grained object distinction. At inference time, natural language queries are matched against the learned feature fields, followed by two-stage 3D clustering to retrieve Gaussian groups corresponding to individual object instances. Unlike prior approaches that depend on 2D rendering-based matching or point-level features, our design enables direct reasoning over 3D feature fields with consistent instance-aware semantics, addressing the limitations in cross-view alignment and object-level coherence identified above.
In summary, the main contributions of this work are as follows:
\begin{itemize}
    \item We introduce \coolname{}, a unified framework for 
    instance-level open-vocabulary 3D scene understanding built on \gls{gs}.
    \item We propose a semantic representation scheme combining a language feature field and an instance-discriminative feature field trained with contrastive supervision.
    \item We develop a two-stage 3D clustering strategy that integrates semantic similarity and spatial coherence to enable accurate instance grouping.
    \item We conduct extensive experiments on standard 3D scene understanding benchmarks, showing that our method achieves competitive or superior performance compared to state-of-the-art open-vocabulary approaches.
\end{itemize}

\section{Related Work}
\label{sec:related}

\subsection{Neural Rendering for 3D Scene Understanding}
\gls{nerf}~\cite{mildenhall2021nerf} introduced a continuous volumetric representation that enables high-quality novel view synthesis, but suffers from slow training convergence and high inference overhead due to dense sampling. 
To address this, \gls{gs}~\cite{kerbl20233d} replaces volumetric rendering with differentiable rasterization of anisotropic Gaussians, achieving a favorable trade-off between speed and fidelity. However, integrating semantic understanding into \gls{gs} remains an open challenge. 

Semantic understanding in 3D aims to enrich reconstructed scenes with region-level annotations. Early work extended \gls{nerf} to encode semantics, often by distilling dense features from pre-trained 2D networks. For instance, Distilled Feature Fields~\cite{kobayashi2022decomposing} integrates features from LSeg~\cite{li2022language} and DINO~\cite{caron2021emerging} into \gls{nerf}, while Semantic NeRF~\cite{zhi2021place} jointly models appearance and semantics for semantic view synthesis. 
In \gls{gs}, Feature3DGS~\cite{zhou2024feature} distills features from SAM~\cite{kirillov2023segment} into Gaussians, while GaussianGrouping~\cite{ye2024gaussian} and ClickGaussian~\cite{choi2024click} incorporate mask grouping and interactive segmentation. Despite their effectiveness, these methods are primarily designed under closed-set supervision and offer limited or no support for open-vocabulary or language-guided understanding.

\subsection{Open-Vocabulary and Language-Driven 3D Representation}
Open-vocabulary 3D representation enables zero-shot concept understanding and free-form language queries in scenes. LERF~\cite{kerr2023lerf} was the first to demonstrate the potential of directly incorporating CLIP~\cite{radford2021learning} embeddings into \gls{nerf}, enabling text-driven queries across scenes. Later work~\cite{liu2023weakly} further added DINO supervision to improve retrieval and segmentation accuracy. While effective, \gls{nerf}-based approaches remain limited by their slow volumetric rendering. 

Recent efforts have begun exploring open-vocabulary understanding within \gls{gs}. LangSplat~\cite{qin2024langsplat} employs a scene-specific language autoencoder to learn latent semantic embeddings, effectively enhancing object boundary delineation in rendered feature maps, while LEGaussians~\cite{shi2024language} attaches uncertainty-aware semantic attributes to Gaussians using quantized CLIP and DINO features.
OpenGaussian~\cite{wu2024opengaussian}, most closely related to our work, links Gaussians to semantic features via a two-stage codebook discretization and introduces a language association to enable point-level open-vocabulary understanding. 
Unlike OpenGaussian, our method does not discretize instance features; instead, it learns a continuous instance feature field and retrieves object instances via density-based clustering in a post-training decoding step.
More recently, Dr. Splat~\cite{jun2025dr} directly registers CLIP embeddings to 3D Gaussians by enforcing global feature alignment through product quantization trained on large-scale image data, demonstrating strong performance in language-guided 3D perception tasks.
While these approaches significantly improve per-Gaussian semantic alignment, they typically rely on quantized features and do not explicitly model instance-level coherence.
Our approach introduces an end-to-end framework that learns view-consistent, instance-aware features over~\gls{gs}, enabling fine-grained object-centric semantic understanding in 3D scenes.

\section{Methodology}
\label{sec:method}

\begin{figure*}[t]
	\centering
	\includegraphics[width=1.0\textwidth]{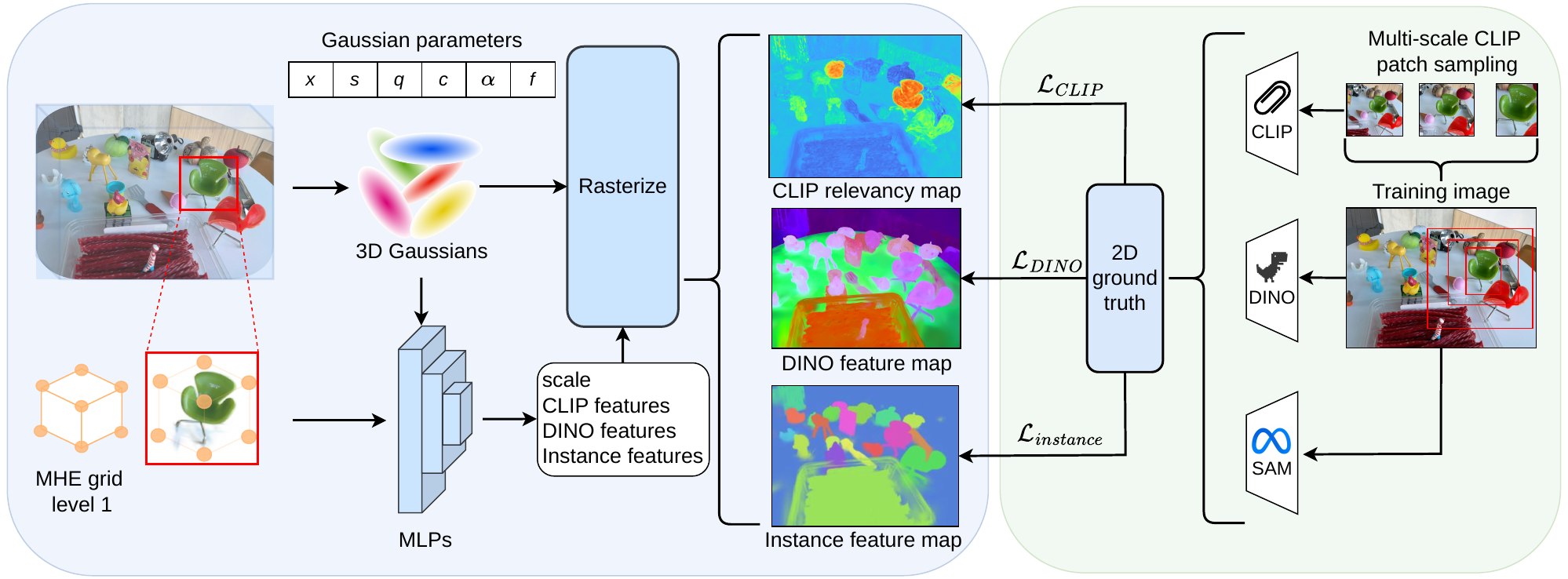}
	\centering
	\caption{An overview of \coolname{}. 
    \textbf{Left}: Our method learns language-aligned and instance-aware features for 3D Gaussians, computed via compact multi-resolution hash encoding and lightweight projection MLPs. 
    \textbf{Right}: Feature learning is guided by multi-view 2D signals, leveraging CLIP for language alignment, DINO for object boundary regularization, and SAM for instance-aware contrastive learning.
    }
	\label{fig:overview}
\end{figure*}

In this section, we present \coolname{}, a novel framework for instance-level open-vocabulary 3D scene understanding built on \gls{gs}. Unlike prior approaches that treat geometry and semantics as separate stages, \coolname{} jointly reconstructs the scene and learns language-aligned, instance-discriminative features within a unified optimization. 
The core technical contribution lies in augmenting each Gaussian splat with learnable semantic attributes, encoded via a \gls{mhe}~\cite{muller2022instant} centered at the Gaussian's position and decoded through lightweight MLPs. This design enables compact 
yet expressive feature representations that scale to dense Gaussian scenes without excessive memory or runtime overhead.
Supervision is guided by a combination of contrastive and pixel-aligned losses derived from 2D foundation models, grounding the learned features in both language semantics and spatial consistency across views. 
As a result, the learned representations become simultaneously language-aligned and instance-discriminative.
Figure~\ref{fig:overview} provides an overview of the full pipeline. We begin by reviewing the fundamentals of \gls{gs} to establish the notation and context for our method. We then detail each component of our approach: the semantic attribute parameterization, the multi-view feature distillation strategy, and the training objectives that bring them together.

\subsection{Preliminary on 3D Gaussian Splatting}
\label{subsec:preliminary}

\gls{gs}~\cite{kerbl20233d} is a neural rendering method that represents a 3D scene as a set of explicit, spatially distributed anisotropic Gaussians. In \gls{gs}, each Gaussian primitive is parameterized by its position $x \in \mathbb{R}^3$, covariance matrix $\Sigma \in \mathbb{R}^{3 \times 3}$, opacity $\alpha \in [0, 1]$, and spherical harmonics (SH) coefficients for modeling appearance. 
The covariance matrix \(\Sigma\) is further decomposed into rotation matrix $R$ and scaling matrix $S$ to ensure that $\Sigma$ is always positive semi-definite for the optimization process. In practice, the rotation matrix $R$
is encoded as a quaternion $q \in \mathbb{R}^4$, and the scaling matrix $S$ as a scaling factor $s \in \mathbb{R}^3$, respectively.

Each Gaussian encodes its color $c$ using spherical harmonics, which enables view-dependent color variations. Rendering is performed via $\alpha$-blending, where the color of a pixel is computed by blending $N$ Gaussians in ascending order of the distances to camera origin (front-to-back):
\begin{equation}
\hat{C} = \sum_{i \in N} c_i \alpha_i \prod_{j=1}^{i-1} (1 - \alpha_j),
\end{equation}
where $\alpha_i$ is the effective opacity of the $i$-th Gaussian. This differentiable formulation ensures that gradients for all parameters, including mean, covariance, opacity, and SH coefficients, are explicitly derived to reduce computational overhead during training.

\subsection{Initialization}
\label{subsec:initialization}

We begin by extracting initial 2D supervision signals from the input RGB images $\{I_v\}_{v=1}^V$, each of resolution $H \times W$ captured from $V$ calibrated camera views. For each image $I_v$, we apply SAM with dense point prompts to produce a set of masks $\{M_{v, k}\}_{k=1}^{K_v}$, where $K_v$ is the number of detected instances in view $v$. SAM produces multiple mask proposals per prompt at different levels of granularity, where we retain only the ``whole'' instance masks~\cite{qin2024langsplat}, as they most reliably capture complete objects across views. Each retained mask is further associated with a CLIP embedding, obtained from the corresponding cropped image patch.
These masks and their associated embeddings are used to construct semantic, pixel-aligned and low-dimensional feature maps for training. 
After initialization, no additional pretrained models are needed during training, aside from the CLIP text encoder used at inference.

To enrich each Gaussian primitive with semantic information, we associate it with one or more learnable, view-independent feature vectors $f_i^t \in \mathbb{R}^{d_t}$, where $f_i^t$  represents a different semantic attribute (e.g., language, instance). These features are projected into 2D views via the \gls{gs} rendering process. For a given feature type $t$, the rendered feature map $\hat{F}^t(p)$ at pixel $p$ is computed as:
\begin{equation}
\hat{F}^t(p) = \sum_{i \in \mathcal{N}(p)} f_i^t \alpha_i \prod_{j=1}^{i-1} (1 - \alpha_j),
\label{eq:feature_render}
\end{equation}
where $\mathcal{N}(p)$ is the ordered set of Gaussians contributing to pixel $p$. This formulation supports end-to-end learning of 3D feature fields from 2D supervision while maintaining differentiability through the Gaussian rasterization pipeline.

\subsection{Language Feature Field}
\label{subsec:language}

Our language feature field is inspired by~\cite{zuo2025fmgs}, which constructs a volumetric embedding over 3D Gaussians using average CLIP features aggregated across multi-scale crops of training images. While effective, this method decouples geometry and semantics by freezing reconstruction before learning semantics, often resulting in misalignment between appearance and language. To address this, we propose a progressive joint optimization of both geometry and semantics, enabling better alignment and consistency across 2D and 3D representations.

To enable efficient language embedding across the set of Gaussian splats in a scene, we parameterize a compact language feature field using a \gls{mhe}~\cite{muller2022instant}. Each Gaussian $G_i$ is associated with a learnable language embedding that is computed as a function of its 3D location $x_i \in \mathbb{R}^3$ and is encoded as $f_i^{lang} = MHE_{\theta}(x_i)$, where $MHE_{\theta}$ denotes the hash-based encoder with parameters $\theta$.
The encoded feature $f_i^{lang}$ is then fed into a lightweight projection MLP to produce the CLIP-aligned embedding $f_i^{CLIP} \in \mathbb{R}^{D_{CLIP}}$ and a scalar $scale_i$ value:
\begin{equation}
f_i^{CLIP}, scale_i = MLP_{\phi}^{CLIP}(f_i^{lang}),
\end{equation}
where $\phi$ are learnable MLP weights and $D_{CLIP}$ is the CLIP feature dimension. The learned $scale_i$ value is used later during semantic clustering in Section~\ref{subsec:clustering} to adaptively group Gaussians at the most relevant level of granularity for a given text query.

To supervise this feature field, we render a 2D language map by projecting Gaussians into camera views and compositing their language features $f_i^{lang}$ via alpha blending (Equation~\ref{eq:feature_render}). While $f_i^{lang}$ is a compact feature vector derived from \gls{mhe} for efficient rasterization, it is subsequently mapped to the full CLIP space through the projection MLP, yielding a predicted language map $\hat{F}^{CLIP}$ with $D_{CLIP}$ dimensional features. We supervise this map using a Huber loss against a hybrid ground-truth CLIP map constructed from multi-scale image crops. This loss encourages each Gaussian to align with multi-view CLIP supervision, thereby enabling the formation of a dense and coherent 3D geometry aligned with natural language.

\subsection{Instance Feature Field}
\label{subsec:instance}

While the language feature field enables open-vocabulary semantic alignment, it does not explicitly distinguish between individual object instances. To address this, we introduce an instance feature field that encodes discriminative embeddings for object-level reasoning. 
Specifically, we associate each 3D Gaussian with a learnable embedding $f_i^{inst}$, optimized via contrastive learning~\cite{choi2024click} over 2D instance masks.

Given an input image $I_v$ from view $v$, and an instance mask $M_{v,k} \subset I_v$ for the $k$-th instance in that view, we learn an 
instance feature map $\hat{F}^{inst}_v$ of dimension $D_{inst}$ by first extracting a representative feature $r_{v,k}$ as the average feature over the corresponding mask:

\begin{equation}
r_{v, k} = \frac{1}{|M_{v,k}|} \sum_{p \in M_{v,k}} \hat{F}^{inst}_v(p).
\end{equation}

A contrastive loss is then applied to enforce that features belonging to the same instance are pulled toward $r_{v, k}$:
\begin{equation}
\mathcal{L}_{pos} = \sum_{v} \sum_{k} \frac{1}{|M_{v,k}|} \sum_{p \in M_{v,k}} \left\| \hat{F}^{inst}_v(p) - r_{v,k} \right\|_2^2.
\end{equation}

To encourage separability across different instances, we apply a margin-based loss on the representative features:
\begin{equation}
\mathcal{L}_{neg} = \sum_{v} \sum_{k \neq k'} \text{ReLU}(\gamma - \|r_{v,k} - r_{v,k'}\|_2^2),
\end{equation}
where $\gamma$ is a margin that defines the minimum allowable distance between distinct instance features.

The total instance contrastive loss becomes:
\begin{equation}
\mathcal{L}_{\text{instance}} =  \mathcal{L}_{\text{pos}} + w_{neg} \cdot \mathcal{L}_{\text{neg}},
\end{equation}
with $w_{neg}$ controlling the trade-off between intra-instance compactness and inter-instance separation.

\subsection{3D Instance Clustering}
\label{subsec:clustering}

Once the instance feature field is trained, we introduce a two-stage clustering strategy that jointly leverages semantic similarity and spatial consistency. 
This clustering process is conducted only once after training converges and serves as the basis for instance-level 3D reasoning. 
Note that clustering in \coolname{} is not used to enforce instance identity as in~\cite{wu2024opengaussian}, but to decode instances from an already instance-aware embedding space.

\subsubsection{Stage 1: Semantic Clustering}
We first perform clustering in the learned instance feature space. 
To enable efficient and unsupervised object grouping, we apply HDBSCAN, a density-based clustering algorithm, over the embedded features $\{f_i^{inst}\}_{i=1}^N$. This step yields a set of initial clusters $\{\mathcal{C}_i\}_{i=1}^{M}$, that group semantically coherent Gaussians. HDBSCAN is well-suited for this task due to its robustness to noise and ability to infer the number of clusters adaptively.

To identify clusters relevant to a text prompt, we compute the average CLIP relevancy for each cluster by comparing the text embedding with the language features $f_i^{lang}$ of its members. Importantly, we adapt this comparison by incorporating the predicted $scale$ values, which effectively control the sensitivity and granularity of cluster relevance.  This idea is inspired by prior work~\cite{kim2024garfield}, but differs in that our formulation operates directly on 3D Gaussians via \gls{mhe}-encoded positions, rather than on pixel-aligned ray samples. Then, clusters with average similarity above a predefined threshold $\tau$ are retained. This relevance filtering step, informed by the predicted scale values, ensures that only semantically aligned and appropriately scaled Gaussian groups are passed to subsequent stages of instance identification.

\subsubsection{Stage 2: Spatial Refinement}
While the first stage captures semantically coherent groups, we found that it may still include spatially disjoint or noisy Gaussians due to the challenges of relying solely on unsupervised clustering and soft CLIP-based similarity.
To address this, we perform a secondary clustering in 3D Cartesian space. For each selected semantic cluster, we apply DBSCAN over the 3D coordinates of its member Gaussians. This step segments each cluster into spatially connected subgroups, and we discard all points marked as noise. 
Importantly, we preserve all valid subclusters to handle cases where multiple instances of the same object category (e.g., multiple ``chairs'') appear in different regions of the scene. This explicitly enables instance-level separation beyond semantic segmentation and ensures robustness to intra-class variance.

\subsection{Training Objectives}
\label{subsec:training}

We jointly train the scene geometry and semantic feature fields by minimizing a combination of photometric and semantic losses. This unified objective encourages both accurate appearance reconstruction and consistent, discriminative feature embeddings across views and instances.

To further enhance spatial consistency and delineate object boundaries, we introduce DINO regularization by applying an auxiliary loss from dense DINO features. This encourages semantic smoothness and improves the localization of object edges, ultimately enhancing the quality of the learned embeddings. The total training loss is formulated as:
\begin{equation}
\mathcal{L}_{\text{total}} = 
\mathcal{L}_{\text{RGB}} + 
\lambda_{\text{lang}} \mathcal{L}_{\text{CLIP}} + 
\lambda_{\text{inst}} \mathcal{L}_{\text{instance}} + 
\lambda_{\text{reg}} \mathcal{L}_{\text{DINO}}.
\end{equation}
Each term is scaled by a weight $\lambda$ controlling its contribution to the overall optimization. This joint optimization enables \coolname{} to achieve accurate reconstruction while learning meaningful and consistent semantic representations across views.

\section{Experimental Design}
\label{sec:exp}

\subsection{Datasets}
We evaluate our method on two benchmark datasets widely used for 3D scene understanding.
For the \textit{3D object selection} task, we evaluate on the LERF dataset~\cite{kerr2023lerf}, which provides multi-view RGB images with open-vocabulary 2D annotations for long-tail objects. This dataset enables evaluation of free-form object queries and projection-based metrics.
For the \textit{3D class-agnostic instance segmentation} task, we adopt the ScanNet dataset~\cite{dai2017scannet}, a large-scale indoor scene benchmark with RGB-D sequences, camera poses, and rich 3D ground truth annotations.

\subsection{Implementation Details}
We implement our method by building upon the \textit{splatfacto} model from Nerfstudio~\cite{tancik2023nerfstudio}, training for 30K iterations on an NVIDIA H100 GPU. The feature field follows the multi-resolution hash encoding with projection MLP design introduced in~\cite{zuo2025fmgs}, with output dimensions of $D_{CLIP} = 512$ and $D_{inst} = 8$, respectively. Optimization is performed using Adam with learning rates of 0.0025 for the 3D Gaussian parameters and 0.001 for the MLPs. Loss weights are set to 0.5 for the CLIP and DINO supervision, and 1.0 for the instance contrastive loss. We use a margin of 1.0 and set the negative pair weight to 0.1 in the contrastive loss. For instance-level clustering, we utilize the CUDA-accelerated HDBSCAN from cuML~\cite{raschka2020machine} with default hyperparameter settings, varying the minimum samples per cluster between 20 and 100 based on scene complexity. 
To ensure a fair comparison, hyperparameters for the comparison baselines follow their original papers or are tuned for best performance on validation scenes.

\subsection{3D Object Selection}
\label{subsec:selection}

\textbf{1) Task:} 
This task evaluates the retrieval and localization of 3D objects in a scene using open-vocabulary text queries. Given a free-form text prompt, we extract a corresponding CLIP embedding and compute cosine similarity with the semantic features assigned to each 3D Gaussian. Gaussian groups with similarity scores above a threshold are selected, and the corresponding 3D points are rendered into 2D views via the \gls{gs} rasterization. This enables direct visual comparison between predicted object masks and 2D ground truth.
\textbf{2) Baseline:} 
We compare our method against recent \gls{gs} based approaches designed for the 3D object selection task, including OpenGaussian~\cite{wu2024opengaussian}, LangSplat~\cite{qin2024langsplat} and LEGaussians~\cite{shi2024language}.
\textbf{3) Metrics:} 
We use the LERF dataset~\cite{kerr2023lerf} with open-vocabulary annotations aligned to rendered 2D images from~\cite{qin2024langsplat}. Performance is measured by \textit{mIoU} and \textit{mAcc} against ground-truth masks.

\subsection{3D Instance Segmentation}
\label{subsec:segmentation}

\textbf{1) Task:}
We evaluate the effectiveness of our instance feature field and clustering strategy in segmenting 3D object instances. The objective is to assign each 3D Gaussian to a unique, class-agnostic instance label, ensuring spatial coherence without relying on predefined semantic classes.
This task differs from point-level semantic understanding, as it requires explicit grouping of Gaussians into object instances rather than per-point relevance estimation.
\textbf{2) Baseline:} 
Our method is compared against recent \gls{gs} based approaches, including OpenGaussian~\cite{wu2024opengaussian} and GaussianGrouping~\cite{ye2024gaussian}, which also leverage RGB images and SAM masks for instance guidance. Moreover, we evaluate Sam3D~\cite{yang2023sam3d}, a strong baseline that directly operates on point clouds with depth input. Unlike Sam3D, our approach relies solely on 3D Gaussians reconstructed from RGB supervision, without using explicit depth.
\textbf{3) Metrics:} 
We conduct an evaluation on the ScanNet~\cite{dai2017scannet} dataset with 3D ground-truth instance annotations. Performance is reported using \textit{mIoU} and \textit{mAcc}.

\section{Results}
\label{sec:Results}

\subsection{3D Object Selection}
\label{subsec:results_selection}

\begin{table}[t]
\centering
\footnotesize
\resizebox{\columnwidth}{!}{%
\begin{tabular}{l|cccc|c|cccc|c}
\toprule
\multirow{2}{*}{\textbf{Methods}} & \multicolumn{5}{c|}{\textbf{mIoU} $\uparrow$} & \multicolumn{5}{c}{\textbf{mAcc.} $\uparrow$} \\
& fig. & tea. & ram. & kit. & Mean & fig. & tea. & ram. & kit. & Mean \\
\midrule
LangSplat*~\cite{qin2024langsplat} & 10.16 & 11.38 & 7.92 & 9.18 & 9.66 & 8.93 & 20.34 & 11.27 & 9.09 & 12.41 \\
LEGaussians*~\cite{shi2024language} & 17.99 & 19.27 & 15.79 & 11.78 & 16.21 & 23.21 & 27.12 & 26.76 & 18.18 & 23.82 \\
OpenGaussian*~\cite{wu2024opengaussian} & 
\underline{39.29} & \underline{60.44} & \textbf{31.01} & \underline{22.70} & \underline{38.36} & \underline{55.36} & \underline{76.27} & \textbf{42.25} & \underline{31.82} & \underline{51.43} \\
\textbf{\coolname{} (Ours)} & \textbf{45.94} & \textbf{62.21} & \underline{21.70} & \textbf{23.95} & \textbf{38.45} & \textbf{64.15} & \textbf{79.17} & \underline{36.36} & \textbf{40.91} & \textbf{55.15} \\
\bottomrule
\end{tabular}
}
\caption{Quantitative comparison of 3D object selection performance on the LERF dataset. (Best and second-best results are highlighted in \textbf{bold} and \underline{underlined}, respectively. Columns correspond to the \emph{figurines}, \emph{teatime}, 
\emph{ramen}, and \emph{waldo kitchen} scenes. Accuracy is measured by mAcc@0.25. *Results reported from~\cite{wu2024opengaussian}).
}
\label{tab:selection}
\end{table}

\begin{figure*}[t]
    \centering
    \begin{minipage}[t]{0.49\textwidth}
        \centering
        \includegraphics[width=\linewidth]{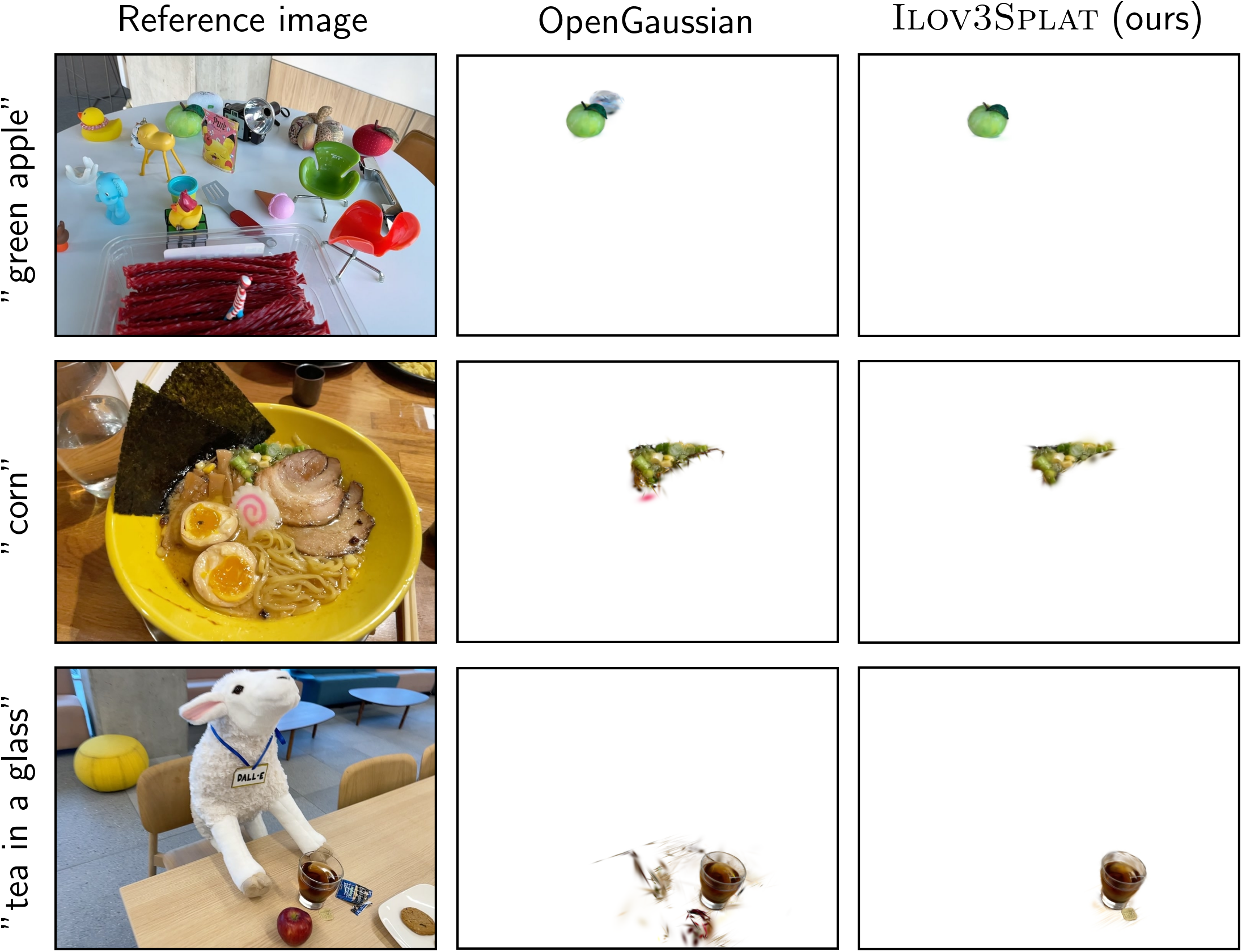}
    \end{minipage}
    \hfill
    \begin{minipage}[t]{0.49\textwidth}
        \centering
        \includegraphics[width=\linewidth]{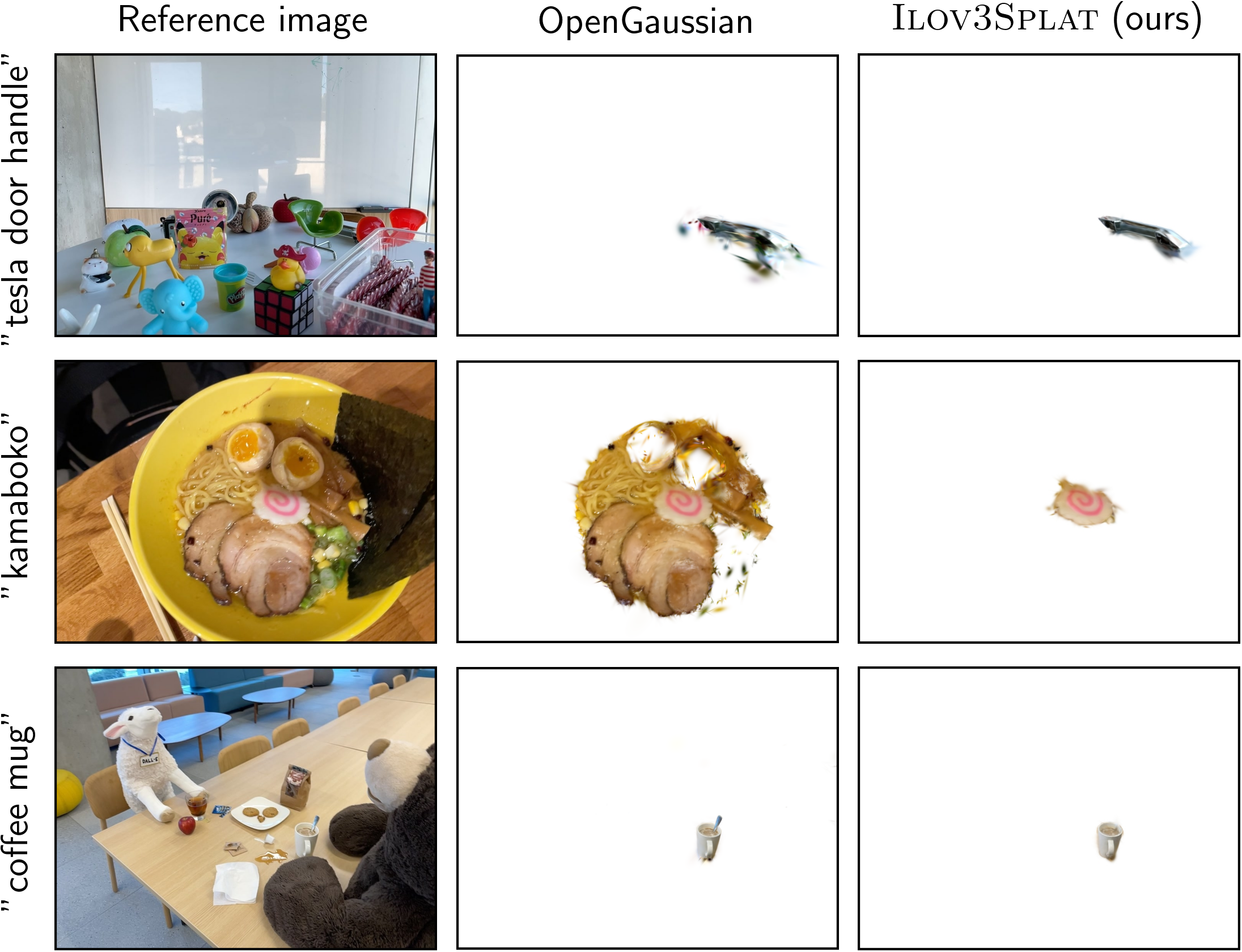}
    \end{minipage}
    \caption{Qualitative results of 3D object selection on the LERF dataset.}
    \label{fig:query_results}
\end{figure*}

Table~\ref{tab:selection} quantitatively compares the 3D object selection performance between our \coolname{} and recent \gls{gs} based baselines on the LERF dataset across four scenes: ``figurines'', ``teatime'', ``ramen'', and ``waldo kitchen''. 
Our method achieves the highest mAcc and a competitive mIoU, outperforming all prior approaches. In particular, \coolname{} consistently ranks first or second across most scenes, demonstrating strong capabilities in both fine-grained semantic retrieval and coherent object boundary delineation.

Qualitative results are illustrated in Figure~\ref{fig:query_results}. OpenGaussian tends to produce false activations due to the lack of instance-level coherence. This results in scattered artifacts in the rendered scenes, as seen in the ``tea in a glass'' example, where unrelated Gaussians are erroneously highlighted. Such artifacts arise because OpenGaussian evaluates points independently, without considering their collective semantic or spatial coherence. In contrast, \coolname{} consistently produces more accurate solutions by leveraging instance-aware feature learning and two-stage 3D clustering. Moreover, our method shows strong resilience to occlusion. In the ``Tesla door handle'' example, despite limited visibility in individual views, \coolname{} successfully localizes the correct object, highlighting its ability to maintain semantic consistency across views.

\subsection{3D Instance Segmentation}
\label{subsec:results_segmentation}

\begin{table}[t]
\centering
\begin{tabular}{l| l| c c}
\toprule
\textbf{Type} & \textbf{Methods} & \textbf{mIoU}~$\uparrow$ & \textbf{mAcc.}~$\uparrow$ \\
\midrule
\multirow{1}{*}{\makecell{With depth}} 
& Sam3D~\cite{yang2023sam3d} & \textbf{34.45} &  \underline{58.24} \\
\midrule
\multirow{3}{*}{\makecell{Without\\ depth}} 
& GaussianGrouping~\cite{ye2024gaussian} & 22.55 & 30.54 \\
& OpenGaussian~\cite{wu2024opengaussian} & 27.32 & 52.44 \\
& \textbf{\coolname{} (Ours)} & \underline{32.08} & \textbf{59.29} \\
\bottomrule
\end{tabular}
\caption{Quantitative comparison of 3D class-agnostic instance segmentation performance on ScanNet dataset. Accuracy is measured by mAcc@0.25. 
}
\label{tab:segmentation}
\end{table}

\begin{figure*}[t]
	\centering
	\includegraphics[width=1.0\textwidth]{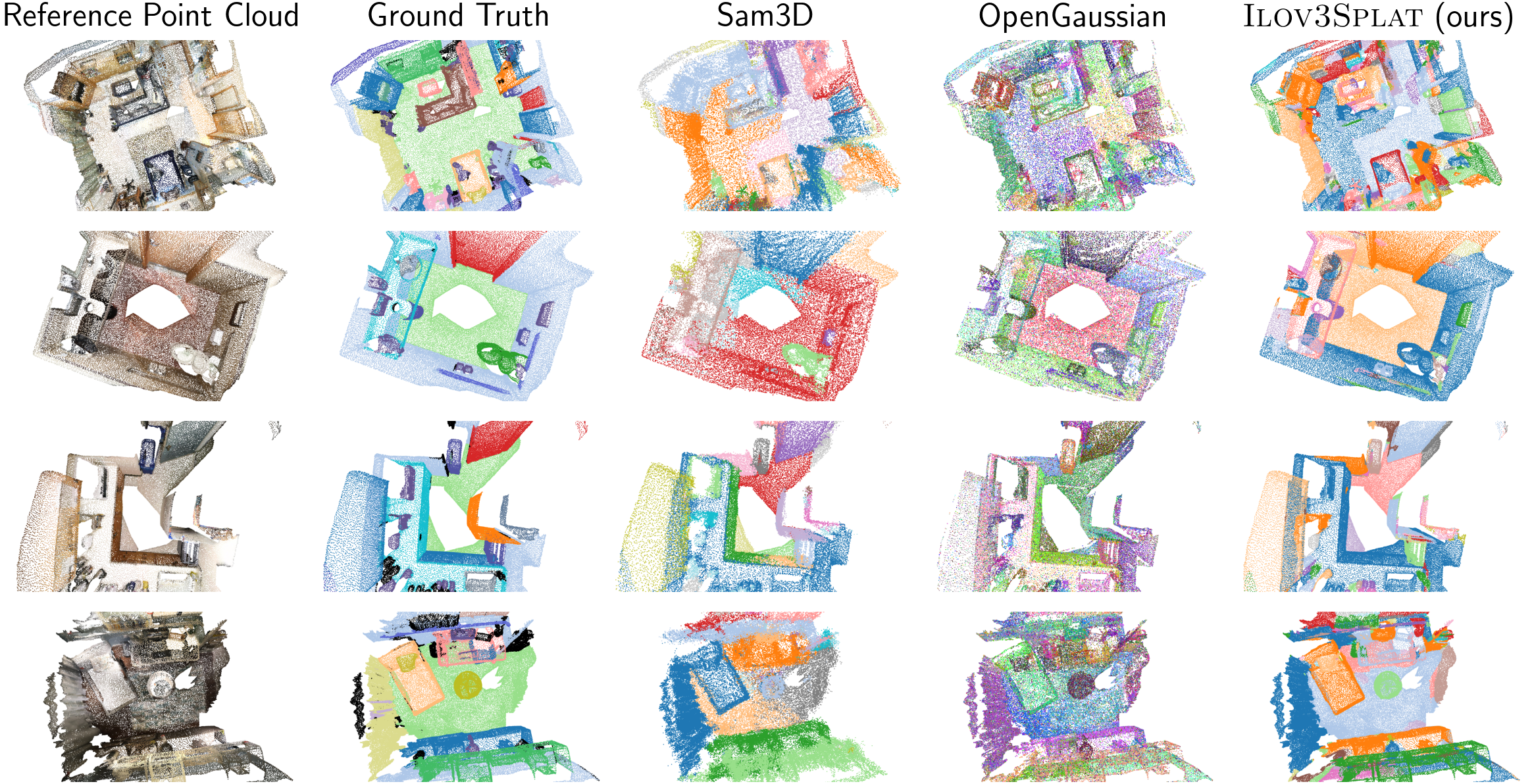}
	\centering
	\caption{Qualitative results of category-agnostic 3D instance segmentation on the ScanNet dataset.}
	\label{fig:segmentation}
\end{figure*} 

The quantitative results on instance segmentation are reported in Table~\ref{tab:segmentation}. 
ScanNet point clouds are notably sparse (approximately 100K points per scene), limiting geometric detail and making it challenging to resolve fine object boundaries. 
In such settings, our method achieves the highest mAcc, outperforming all baselines including Sam3D, which benefits from depth information. In terms of mIoU, our method remains competitive, outperforming all Gaussian-based approaches and closely approaching Sam3D. 

To further understand these results, we provide a qualitative visualization in Figure~\ref{fig:segmentation}. 
Sam3D segments the scene by projecting 2D masks into 3D using depth information and merges adjacent regions using a bidirectional overlap grouping. This often results in coarse segmentations where the scene is divided into large, indistinct regions. 
In contrast, OpenGaussian applies point-level semantic clustering, which leads to noisy segmentations, often failing to separate fine objects. 
\coolname{} strikes a balance between these extremes: it avoids excessive merging while still maintaining coherent and compact instance masks, resulting in more accurate segmentation of both large structures and small furniture items.

\subsection{Ablation Study}

\begin{table}[t]
\centering
\begin{tabular}{l ccc cc}
\toprule
\textbf{Configuration} & \textbf{DINO} & \textbf{Stage 1} & \textbf{Stage 2} & \textbf{mIoU} $\uparrow$ & \textbf{mAcc.} $\uparrow$ \\
\midrule
\textbf{Training components}\\
\multicolumn{4}{l}{Without \gls{mhe} }  & 22.65 & 34.49 \\
\multicolumn{4}{l}{Without joint optimization}  & 36.71 & 45.22 \\
\midrule
\textbf{Pipeline components} \\
DINO  & \checkmark & -- & -- & 10.97 & 18.88 \\
DINO + S1  & \checkmark & \checkmark & -- & 28.52 & 49.44 \\
DINO + S1 + S2  & \checkmark & \checkmark & \checkmark & \textbf{38.45} & \textbf{55.15} \\
\bottomrule
\end{tabular}
\caption{Ablation of \coolname{}. \textbf{Training components}: training without joint optimization and \gls{mhe}. \textbf{Pipeline components}: DINO regularization, semantic grouping, and spatial refinement. The full model achieves the best mIoU and mAcc.}
\label{tab:ablation}
\end{table}

\begin{figure}[t]
    \centering
    \begin{subfigure}{0.48\linewidth}
        \centering
        \includegraphics[width=\linewidth]{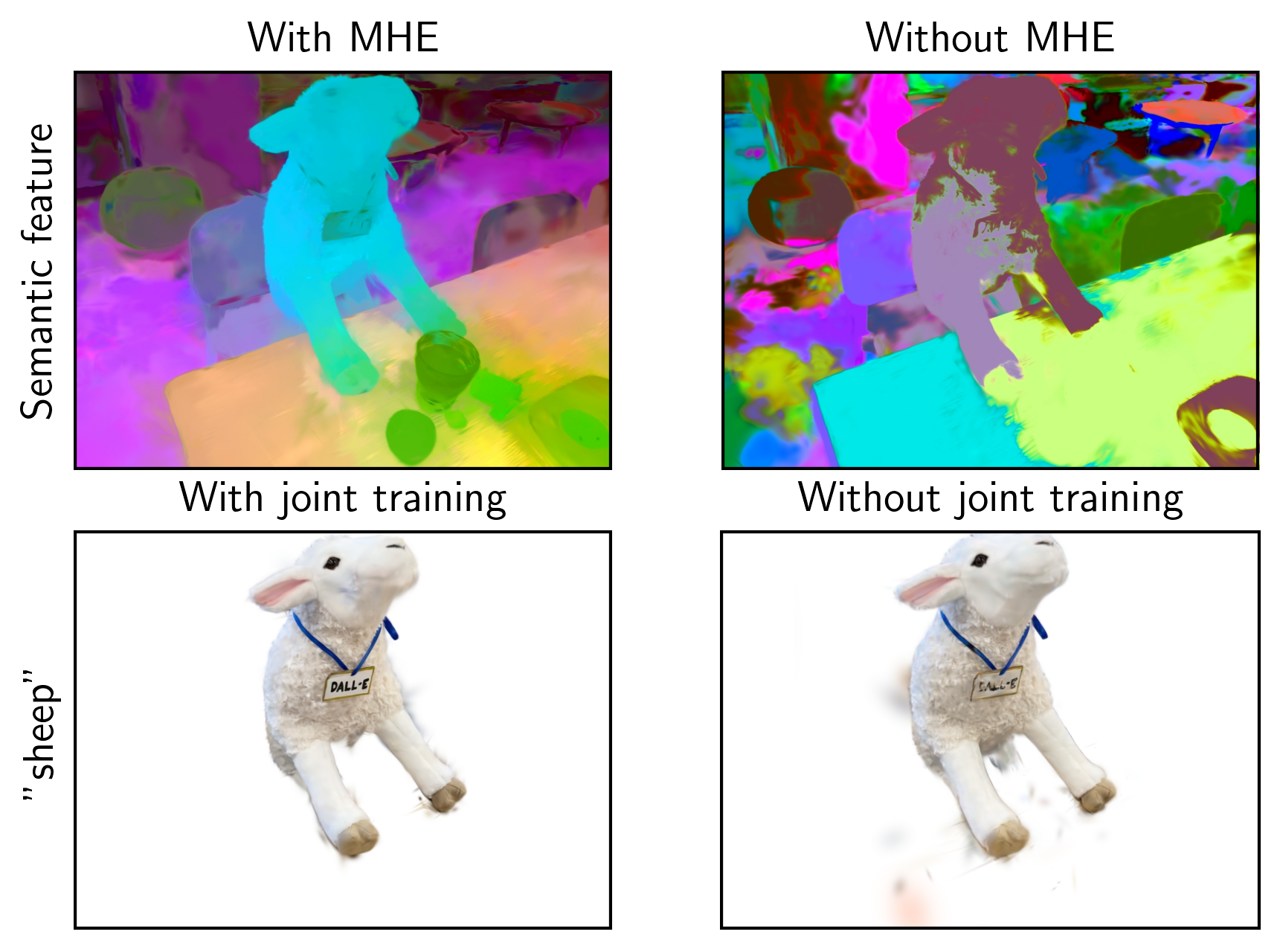}
        \caption{Ablation on training components}
        \label{fig:ablation_a}
    \end{subfigure}\hfill
    \begin{subfigure}{0.48\linewidth}
        \centering
        \includegraphics[width=\linewidth]{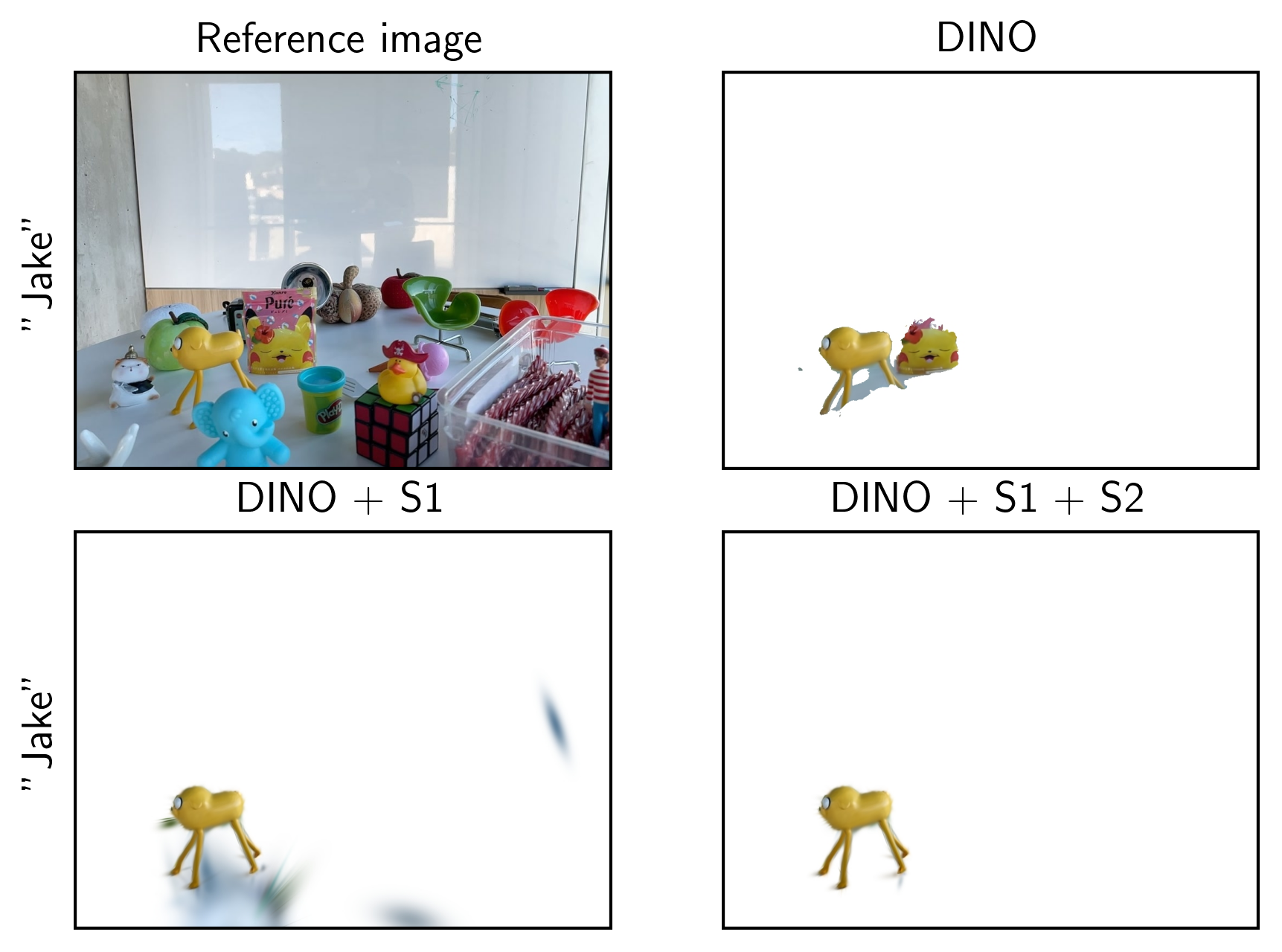}
        \caption{Ablation on pipeline components}
        \label{fig:ablation_b}
    \end{subfigure}
    \caption{Qualitative ablation of \coolname{}. We visualize the impact of individual training components and pipeline stages in our model.}
    \label{fig:ablation}
\end{figure}

\subsubsection{Effect of training components.} An ablation study is conducted to assess the contributions of key training components in \coolname{}. As shown in Table~\ref{tab:ablation}, removing joint optimization or \gls{mhe} leads to noticeable performance degradation. The \gls{mhe} provides a compact and spatially regularized representation for semantic features, which stabilizes optimization and prevents representation collapse. Joint optimization, on the other hand, mitigates appearance–language misalignment by allowing semantic features to co-evolve with the underlying geometry, resulting in more coherent and instance-consistent embeddings.

\subsubsection{Effect of pipeline components.} We observe consistent gains from progressively adding pipeline stages. Starting with DINO-based boundary regularization, incorporating semantic grouping significantly improves both accuracy and coherence by enforcing feature compactness within object-level regions. Adding the spatial refinement stage further enhances performance by filtering spatial outliers and isolating cohesive object parts. Qualitative results in Figure~\ref{fig:ablation} corroborate these quantitative improvements.

\section{Conclusion}
\label{sec:conclusion}
 
We introduce \coolname{}, a unified framework for instance-level open-vocabulary 3D scene understanding based on \gls{gs}. Our method augments Gaussians with compact, view-consistent language and instance feature fields for fine-grained, language-driven reasoning. Through contrastive learning and a two-stage clustering strategy, \coolname{} enables accurate object retrieval and segmentation without relying on predefined categories. Extensive experiments demonstrate competitive or superior performance over existing methods. 
While effective, our method faces challenges in sparse scenes with large objects, where inconsistent multi-view mask supervision can hinder unified instance representations.
In future work, we plan to explore more refined instance feature learning strategies and extend our framework to dynamic environments.

\subsubsection{Acknowledgements} This work was supported in part by the Australian Research Council Discovery Project under Grant DP250103634, and in part by the Commonwealth Scientific and Industrial Research Organisation (CSIRO). The authors acknowledge continued support from the CSIRO’s Embodied AI Cluster.

\bibliographystyle{splncs04}
\bibliography{mybibliography}

\end{document}